\documentclass{article}
\usepackage{ucs}

     \usepackage[final]{neurips_2019}

\usepackage[utf8]{inputenc} 
\usepackage{textcomp}
\usepackage[T1]{fontenc}    
\usepackage{hyperref}       
\usepackage{url}            
\usepackage{booktabs}       
\usepackage{amsfonts}       
\usepackage{nicefrac}       
\usepackage{microtype}      
\usepackage{amsmath}
\usepackage{natbib}
\PassOptionsToPackage{options}{natbib}
\setcitestyle{numbers}
\setcitestyle{square}
\setcitestyle{comma}
\usepackage{lipsum}
\usepackage{graphicx}
\usepackage{epsfig}
\usepackage{subcaption}
\usepackage[ruled,vlined]{algorithm2e}
\usepackage[justification=centering]{caption}
\usepackage{amsmath}

\DeclareMathOperator*{\argmin}{arg\,min}
\title{Tracing State-Level Obesity Prevalence from Sentence Embeddings of Tweets: A Feasibility Study}

\author{%
  Xiaoyi Zhang \\
  Center for Data Science\\
  New York University\\
  \texttt{xiaoyi.zhang@nyu.edu} \\
   \And
   Rodoniki Athanasiadou \\
   School of Medicine \\
   New York University \\
   \texttt{rodoniki.athanasiadou@nyulangone.org} \\
   \AND
   Narges Razavian \\
   School of Medicine \\
   New York University \\
   \texttt{narges.razavian@nyulangone.org } \\
}
\begin{document}
\maketitle
\begin{abstract}
  Twitter data has been shown broadly applicable for public health surveillance. Previous public health studies based on Twitter data have largely relied on keyword-matching or topic models for clustering relevant tweets. However, both methods suffer from the short-length of texts and unpredictable noise that naturally occurs in user-generated contexts. In response, we introduce a deep learning approach that uses hashtags as a form of supervision and learns tweet embeddings for extracting informative textual features. In this case study, we address the specific task of estimating state-level obesity from dietary-related textual features. Our approach yields an estimation that strongly correlates the textual features to government data and outperforms the keyword-matching baseline. The results also demonstrate the potential of discovering risk factors using the textual features. This method is general-purpose and can be applied to a wide range of Twitter-based public health studies.
\end{abstract}

\section{Introduction}
Twitter, or social media in general, is a vast space for users to express opinions and sentiments. The proliferation of social media networks accelerates the generation of public health data at an unprecedented rate, allowing big data computing approaches to achieve innovative and impactful research in health sciences. Since \citet{paul2011you} proposed the use of Twitter data for public health informatics, several studies \citep{zou2016infectious,abbar2015you,nguyen2016building,nguyen2017twitter, sarma2019estimating} discovered strong correlations between government statistics for specific diseases and tweets on specific topics. These studies suggest that a large number of relevant tweets can provide insight into the general health of a population. \\
\hspace*{\fill}\\
In a recent literature review, \citet{jordan2019using} showed the majority of Twitter studies in the last decade on public health informatics has relied on keywords for classification or clustering. However, since tweets contain unpredictable noises such as slang, emoji, and misspellings, the tweets retrieved from keyword-matching models often exclude relevant or include irrelevant messages. These uncertain semantics of retrieved tweets lead to unreliable estimations for public health metrics \citep{jordan2019using, culotta2010towards}. Meanwhile, other studies use Latent Dirichlet allocation to learn latent topics \citep{paul2014discovering, prier2011identifying}. These models depend on reliable word co-occurrence statistics and typically suffer from data sparsity when applied to short documents like tweets \citep{sridhar2015unsupervised}. \\
\hspace*{\fill}\\
We propose using sentence embeddings by supervised deep learning methods to overcome this shortcoming. Hashtags, the user-annotated label that clusters tweets with shared topics regardless of the diverse textual patterns, provide a natural supervision for training distributed representations of tweets. In this work, we adapt \textit{TagSpace} \citep{weston2014tagspace}, a convolutional neural network (CNN) that learns word and tweet embeddings in the same vector space using hashtags as supervised signals.
In order to demonstrate the feasibility of this method, we address the specific task of estimating state-level obesity from tweets characterizing actual dietary habits.
We use both embeddings to cluster and to extract relevant textual features that correspond to population-level dietary habits from over two hundred million tweets. The regression on these textual features strongly correlates to state-level obesity prevalence surveyed by Centers for Disease Control and Prevention \citep{cdc}.
Since our method is not specifically tailored to obesity research, our approach is applicable to a wide range of public health studies that involve Twitter data.

\section{Data acquisition and pre-processing}
\begin{figure}
  \centering
  \captionsetup{type=figure}
  \includegraphics[scale=0.6]{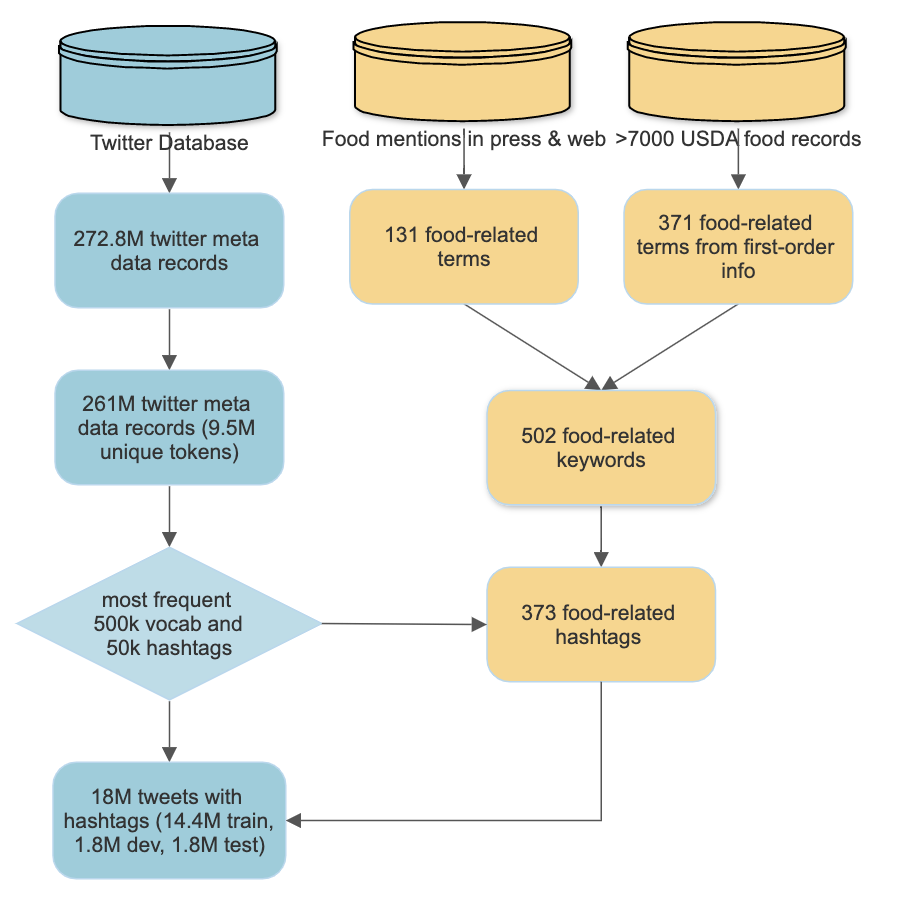}
  \caption{Twitter data pre-processing and keyword acquisition pipeline.}
  \label{pipeline}
\end{figure}

We retrieve 272.8 million tweet records posted in 2014 using the Twitter API \footnote{https://developer.twitter.com/}, and we assign one state among the contiguous United States (48 states plus the District of Columbia) to each of the 261 million records based on user geolocation metadata. Non-English posts are removed from our dataset. 
We use regex to perform the following steps:
\begin{enumerate}
    \item Convert all alphabetical characters to lower case
    \item Remove all URLs, user mentions, and special characters except the hashtag symbol \texttt{\#}
    \item Remove numerical characters except those in hashtags
    \item Add white space between consecutive emojis, and limit repeating mentions of words or emojis in a post
\end{enumerate}
After initial preprocessing, we find 9.5 million unique vocabularies (including hashtags and emojis) heavily tailed at scarce mentions - 6.3 million are mentioned only once, and 8.9 million less than ten times. Similar distribution is also discovered in hashtags. Including scarcely-mentioned words not only results in a memory-demanding lookup table, but also puts our model under the risk of overfitting, as the model may memorize the rare textual patterns found only in the training set. Hence we select 500k most mentioned words (excluding stop words) and 50k hashtags for our model, and all out-of-vocabulary words are tokenized as <UNKNOWN>. The data pre-processing pipeline can be visualized in Figure \ref{pipeline}.
\section{Methods}
We address the specific task of using the twitter data within a state to estimate the obesity prevalence in that state. Inspired by recent Twitter-derived public health studies \citep{zou2016infectious, nguyen2017twitter, nguyen2016building, abbar2015you}, we first compile a set of keywords related to dietary habits to form a feature space in a regression scenario. We then adapt two deep learning models to retrieve food-related tweets by the scoring between embeddings of tweets and keywords. Embeddings of food-related tweets within a state will be aggregated for extracting features later used in regression. 
\subsection{Constructing feature space from keywords}
Following prior works by \citet{nguyen2017twitter}, we generate the keyword list from two sources: 1) the U.S. Department of Agriculture’s National Nutrient Database \citep{cdcfood} - from over 7000 food records found in the USDA database, we extract only the first-level information (e.g. "strawberry yogurt" and "nonfat yogurt" are both recorded as "yogurt"), which gives us 371 terms; and 2) popular food-related mentions in the press \footnote{App Spring Inc. List Challenges: Food, https://www.listchallenges.com/lists/food} - we add food (e.g. "sashimi" and "kimchi"), food-related slangs (e.g. "blt"), and chain restaurants (e.g. KFC and Starbucks) that are not included by the USDA database but frequently appear in user-generated contexts, which results in 131 additional terms. All of the 502 keywords in the list are reduced to their singular forms by NLTK lemmatizer \citep{nltk}, and words in the tweets are also lemmatized for keyword matching. We show the keyword acquisition process in Figure \ref{pipeline}.
\subsection{Retrieving relevant tweets via deep learning}
\begin{figure}
  \centering
  \captionsetup{type=figure}
  \includegraphics[scale=0.42]{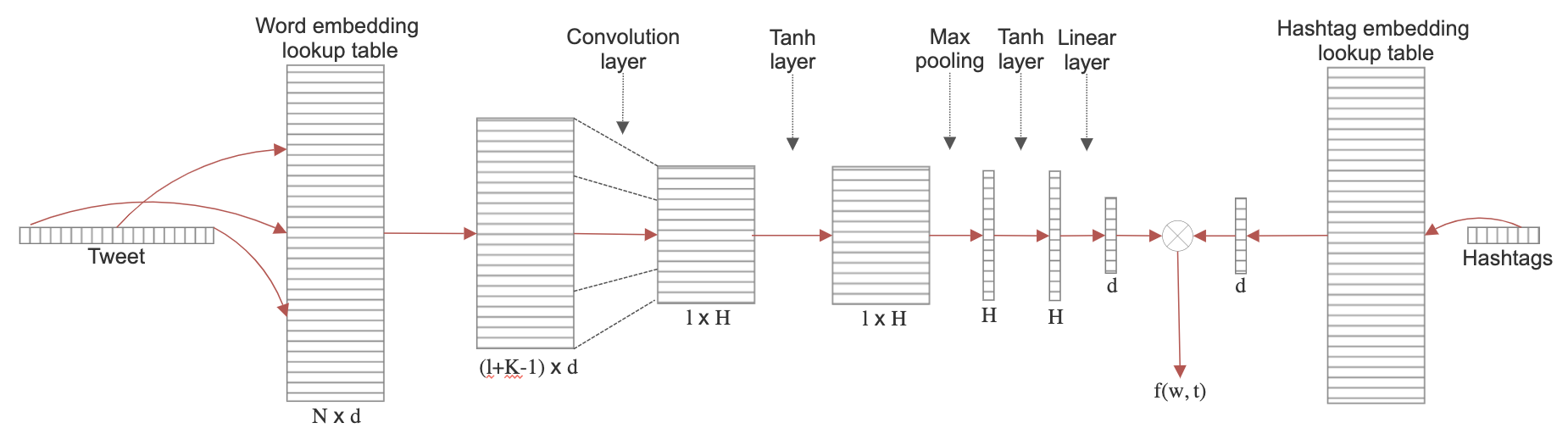}
  \caption{The model architecture of \textit{TagSpace}\citep{weston2014tagspace}. Given an input tweet $w$ and its hashtag $t$, the forward pass outputs the scoring $f(w,t)$, where $N$ denotes vocabulary size, $d$ denotes embedding dimension, $l$ denotes number of words (i.e. max sequence length) of input tweet, $K$ denotes convolution window size, $H$ denotes hidden dimension.}
  \label{forward}
\end{figure}
\paragraph{Keyword matching} The simple baseline regards tweets explicitly mentioning at the one of the keywords as relevant. We find the relative frequency of food-related tweets ranges from 3.0\% to 6.2\% across all states with the mean value of 4.7\%, which is close to statistics reported by \citet{nguyen2017twitter} and \citet{ghosh2013we}. This implies that our pre-processing pipeline, which set a higher bar for word frequency than \citep{nguyen2017twitter} and \citep{ghosh2013we}, does not drastically change the distribution of food-related tweets.
\paragraph{TagSpace} Simple keyword matching results in questionable semantic relevancy of retrieved tweets (e.g. "that problem is a hard \textit{nut} to crack", "taylor swift is the \textit{cream} of the crop"). In contrast, hashtags provide the user's labeling of a tweet's themes. We adapt \textit{TagSpace}, a CNN model that learns the distributed representations of both words and tweets using hashtags as a supervised signal \citep{weston2014tagspace}. Given a tweet, the model convolves its unigrams' embeddings as input, and ranks the scoring (e.g. inner product) between the learnt embeddings of the tweet and candidate hashtags. The ranking is optimized by a pairwise hinge objective function as given by Algorithm \ref{warp}, which is optimized for retrieving top-ranked hashtags according to \citet{weston2014tagspace}. We show the model architecture in Figure \ref{forward}. In our study, we consider a tweet with top ranked hashtags containing the keywords as food-related. Among the 50k hashtag candidates, 373 are included in our food keyword list. For this training task, we require the input tweets to mention hashtags in the candidate pool for prediction, and this gives us 14.4 million tweets for training, 1.8 million held out for hyperparameter tuning and 1.8 million for testing 
To prevent data leakage, hashtags in the input texts are substituted with the corresponding plain words. 
\begin{algorithm}[H]
\SetAlgoLined
\KwData{$e_{conv}(w) \in \mathbb{R}^{N\times d}$: sentence embedding of twitter $w$; $e(t)\in \mathbb{R}^d$: word embedding of hashtag $t$; $T$: set of all hashtags in corpus; $T^+$: set of hashtags found in $w$; $m \in \mathbb{R}$: margin; $M$: max sample iterations}
\KwResult{Optimized CNN weights, $e_{conv}(w)$, and embedding of words in $w$}
Sample $t^+ \in \mathbb{R}^n$ from $T^+$\\
Compute $f(w, t^+) = e_{conv}(w) \cdot e(t^+)$\\
Sample $t^-$ from $T\backslash T^+$\\
Compute $f(w, t^-) = e_{conv}(w) \cdot e(t^-)$\\
Initialize $i \leftarrow 0$\\
 \While{$f(t^-, w) \leq m + f(t^+, w)$ and $i \leq M$} {
   Resample $t^-$ from $T\backslash T^+$\;
   Compute $f(w, t^-) = e_{conv}(w) \cdot e(t^-)$ \;
   $i \leftarrow i + 1$
   }
Compute $Loss= max(0, m-f(w, t^+)+f(w, t^-))$\\
Backward propagation on $Loss$
 \caption{WARP ranking loss of \textit{TagSpace}}
 \label{warp}
\end{algorithm}
\paragraph{Binary TagSpace} While \citet{weston2014tagspace} optimizes the prediction of $p(hashtag\ |\ tweet)$ over all hashtag candidates, we are only interested in the tweets' semantic relevancy with food (i.e. $p(hashtags\ about\ food\ |\ tweet)$). Based on whether a tweet contains hashtags found in our food keyword list, we label all tweets with hashtags as either food-related or not. The word and tweet embeddings learnt from the CNN discussed in the previous method are optimized for a binary classification objective function instead. We use the same training and testing sets as those of the previous method.

\subsection{Feature engineering and obesity estimation by elastic net}
For a given state, features are calculated from the scoring (e.g. inner product or cosine similarity) between the keyword embeddings and the average sentence-level embedding of food-related tweets within that state. The scoring function is the same for both CNN models. Both CNNs' objective functions internally train tweet embeddings in the word vector space \citep{weston2014tagspace, wu2018starspace}, and hence the scoring provides information about the semantic relevance between the tweets and the keywords. By aggregating food-related tweets (i.e. tweets with sentence embeddings that have a high score with keyword vectors) within a state, we represent the dietary characteristics of that state in the word vector space. For obesity prevalence estimation, we apply the elastic net, a regression method that combines L1 and L2 regularization and has been shown to surpass ridge or lasso regressions in text regression tasks \citep{zou2016infectious}. In particular, given the regression task
$$y_s = \textbf{w}^T\textbf{x}_s + \beta + \varepsilon$$
where $y_s \in \mathbb{R}$ denotes the obesity prevalence of a given state $s$, $x_s \in \mathbb{R}^{373}$ the vector of extracted textual features of state $s$, $\beta \in \mathbb{R}$ the intercept, and $\varepsilon \in \mathbb{R}$ an independent and zero-centered noise, the weight vector $w$ is learnt by optimizing the objective function
$$\argmin_{\textbf{w}, \beta}\left(\sum_{s\in S}( \textbf{w}^T\textbf{x}_s + \beta + \varepsilon - y_s)^2 + \lambda_1\sum_{k=1}^{373} |w_k|+ \lambda_2\sum_{k=1}^{373}|w_k|^2  \right) $$
where $\lambda_1$ and $\lambda_2$ are the regularization coefficients, and in practice they are chosen by random search in the range $[1e-5, 1e2]$. 
We randomly hold out four states for validation and eight states for testing, and apply cross-validations for training.
\section{Results and discussion}
\subsection{Deep learning model performance}
\begin{table}[!htb]
    \caption{Performance of CNNs on the test set}
    \begin{subtable}{.5\linewidth}
      \caption{Ranking by \textit{TagSpace}}
      \centering
 \begin{tabular}{ccc}
    \toprule
    Embedding dim & Precision@1 & Recall@10\\
    \midrule
    64 & 15.37\%&40.13\% \\
    128& 28.39\%&43.97\%\\
    256&\textbf{32.72\%}&\textbf{45.65\%}\\
    \bottomrule
    \end{tabular}
    \end{subtable}%
    \begin{subtable}{.5\linewidth}
      \centering
        \caption{Classification by \textit{Binary TagSpace}}
        \begin{tabular}{ccc}
           \toprule
        Embedding dim & Precision & Recall\\
        \midrule
        64 &73.05\%&53.24\% \\
        128 &84.35\%&62.14\% \\
        256 &\textbf{87.48\%}&\textbf{66.37\%}\\
        \bottomrule
        \end{tabular}
    \end{subtable} 
\end{table}
We show the performance of two deep learning models in Table 1 based on their objective functions. Table 1a evaluates the ranking performance of our adaption of \textit{TagSpace}, and the result is comparable to the implementation by \citet{weston2014tagspace} on less noisy text data, which yield 37.42\% P@1 and 43.01\% R@10. This implies that \textit{TagSpace} maintains its ability to predict hashtags on short and noisy documents and hence applicable to Twitter texts in general. As for the binary version of \textit{TagSpace} shown in Table 1b, there is no prior studies for comparison. The low recall can be explained by the unbalanced labels, as in average only 9.4\% of tweets in the test set contain food-related hashtags. The precision of binary \textit{TagSpace} is high, and hence we suspect if the model optimizes objective function by over-generalizing hashtag predictions (i.e. tagging tweets with only general and frequent hashtags such as \texttt{\#restaurant} and \texttt{\#diet}). As the model internally learns tweet embeddings, we use them to rank hashtags and find that the most frequent 100 food-related hashtags in the prediction account for 11.3\% of the food-related tweets. This implies that binary \textit{TagSpace} gives more granular information about a tweet than whether food-related or not. 
\subsection{Estimating obesity prevalence by tweet embeddings}
\begin{table}[!htb]
    \caption{Regressions on obesity prevalence by extracting features from word and tweet embeddings}
      \centering
 \begin{tabular}{cccc}
    \toprule
    Model & dim & MAE & Pearson Corr. \\
    \midrule
    Bag-of-Words & -&2.596 & 0.607 \\
    \midrule
    \phantom{-}&64 & 1.653&0.795 \\
    \textit{TagSpace}&128&1.571&0.813\\
    \phantom{-}&256&1.452&0.836\\
    \midrule
    \phantom{-}&64 & 1.239&0.871 \\
    \textit{Binary TagSpace}&128& 1.018&0.904\\
    \phantom{-}&256&\textbf{0.839}&\textbf{0.927}\\
    \bottomrule
    \end{tabular}
\end{table}
We evaluate the regression results using mean absolute error (MAE) and Pearson correlation with government obesity data. Since no prior study has used our dataset, we handcraft a \textit{Bag-of-Words} baseline that uses tweets filtered by \textit{keyword matching} method and extracts features by frequencies of keywords mentioned within a state. The BOW approach is used in previous Twitter-derived obesity research \citep{nguyen2017twitter, nguyen2016building}. The regression results by our baseline moderately correlates to government data, which agrees with prior works that that dietary characteristics mined from Twitter data is informative in actual obesity estimation \citep{nguyen2017twitter,nguyen2016building,abbar2015you}. Both CNNs generate features resulting in more accurate estimation of state-level obesity compared to our baseline, and binary \textit{TagSpace} outperforms all other methods. Hence we are optimistic that word and tweet embeddings trained from \textit{TagSpace} models optimizing for selective topics results in better indicators of specific diseases.
\subsection{Discovering dietary risk factors with obesity}
\begin{center}
\captionsetup{type=figure}
  \includegraphics[scale=0.22]{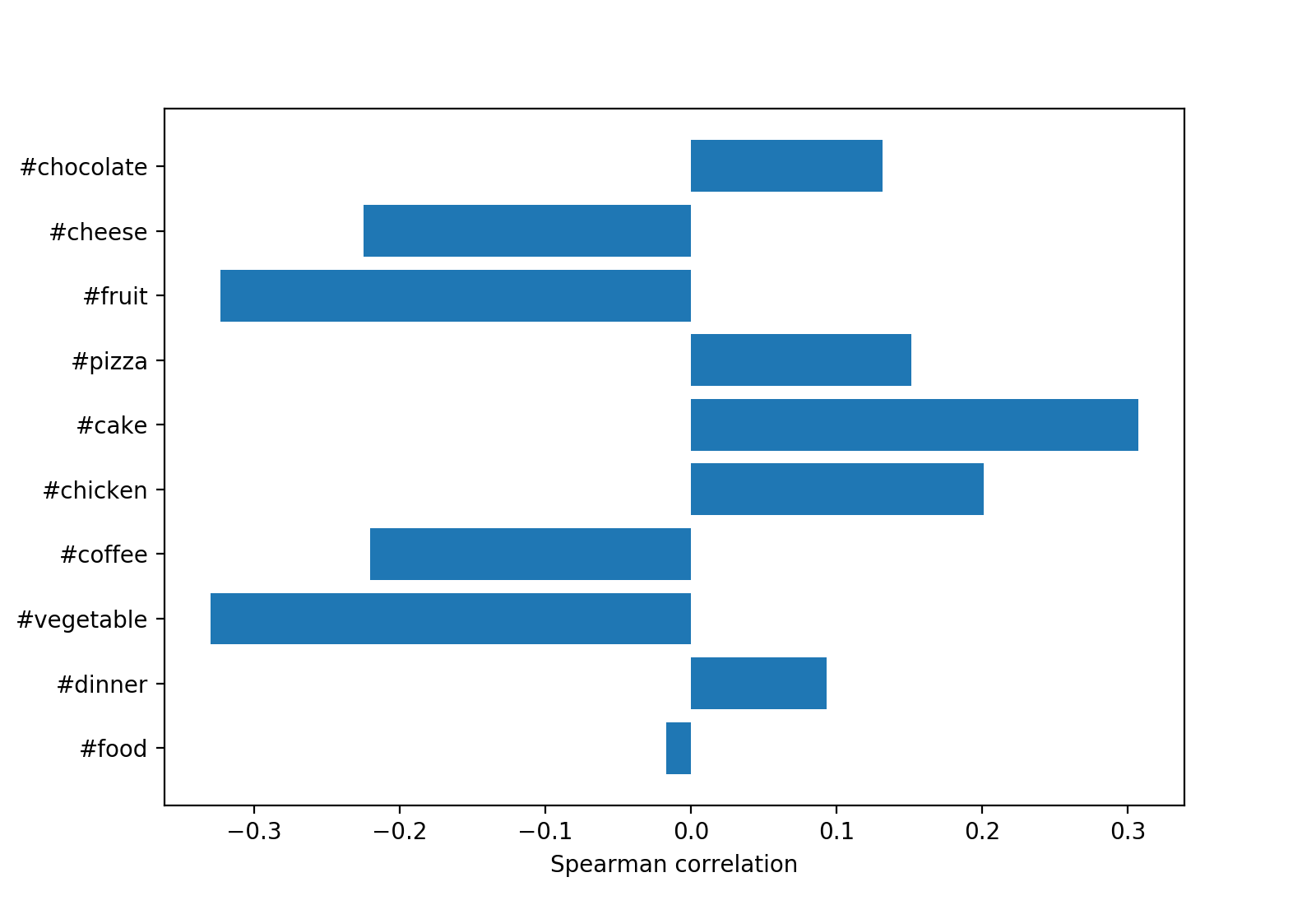}
  \captionof{figure}{Spearman correlations between selected features and obesity prevalence}
  \label{foodcorr}
\end{center}
We are interested in features correlating to higher obesity prevalence, and we obtain such features using Spearman correlation, which quantifies monotonic relationship between two variables. The highest positive correlation to obesity prevalence is given by \texttt{"\#macncheese"} (\textit{corr} = 0.4910), \texttt{"\#wendys"} (0.4853), \texttt{"\#doughnut"} (0.4796), \texttt{"\#blt"} (0.4359), and \texttt{"\#dominospizza"} (0.4307). We also observe that more general and frequently-mentioned features (such as \texttt{\#dinner}, \texttt{\#food}) usually have weaker monotonic relationship with our target variable as shown in Figure \ref{foodcorr}. While the correlation values are modest, it points to a possibility to learn behavior risk factors from Twitter data using sentence-level embeddings of tweets.
\section{Conclusion}
In conclusion, we propose a deep learning approach to extract textual features using sentence-level embeddings of tweets for public health monitoring. In the case study, our adaption of the two CNNs perform reliably on Twitter data and provides informative textual features for obesity prevalence estimation.We have also showed that features constructed via word and tweet embeddings can potentially learn risk factors for specific diseases, which is useful for monitoring acute public health incidents such as influenza tracking, allergy ailments, and infectious diseases \citep{paul2011you, paul2014discovering, zou2016infectious}.
Our data acquisition and deep learning methods do not include any obesity-related settings, which implies that our approach can be applied to a wide range of Twitter-based public health studies and for various purposes. One limitation of our study is that the demographics of Twitter users over-represent younger age-groups, and one remedy is to standardize tweets based on user ages inferred from probabilistic models \citep{chamberlain2017probabilistic} for future work.
We hope this work will inspire future studies to explore the potential of using sentence-level embeddings of social media texts for a wide scope of public health surveillance.

\newpage
\medskip
\small
\bibliography{paper}
\bibliographystyle{plainnat}

\end{document}